# Medical Image Enhancement Using Histogram Processing and Feature Extraction for Cancer Classification

Sakshi Patel
Dept. of Communication Engeneering,
Vellore Institute of Technology,
Tamil Nadu, India
thesakshipatel@gmail.com

Bharath K P
Dept. of Communication Engeneering,
Vellore Institute of Technology,
Tamil Nadu, India
bharathkp25@gmail.com

Rajesh Kumar Muthu
Dept. of ECE,
Vellore Institute of Technology,
Tamil Nadu, India
mrajeshkumar@vit.ac.in

*Abstract*—MRI (Magnetic Resonance Imaging) is a technique used to analyze and diagnose the problem defined by images like cancer or tumor in a brain. Physicians require good contrast images for better treatment purpose as it contains maximum information of the disease. MRI images are low contrast images which make diagnoses difficult; hence better localization of image pixels is required. Histogram Equalization techniques help to enhance the image so that it gives an improved visual quality and a well defined problem. The contrast and brightness is enhanced in such a way that it does not lose its original information and the brightness is preserved. We compare the different equalization techniques in this paper; the techniques are critically studied and elaborated. They are also tabulated to compare various parameters present in the image. In addition we have also segmented and extracted the tumor part out of the brain using K-means algorithm. For classification and feature extraction the method used is Support Vector Machine (SVM). The main goal of this research work is to help the medical field with a light of image processing.

## I. INTRODUCTION

Signal Processing is a vast area of research consisting of various fields, one among them are, the Digital Image Processing (DIP) [1] which allow us to play with components of images as required in desired application. DIP has vast area of research and is used in various fields such as medical imaging, satellite images of planets, and also many industrial applications. Among all these applications, medical field mostly depends on images such as MRI, X-rays, Ultrasound, CT scan and other bio-medical images to identify the exact problem in the patient's body. These images give the detail study of various diseases such as brain tumor, cancer, swelling, etc. So for the physicians to treat and diagnose the problem in a better way needs the images to be in good quality giving all the necessary information about the infected body part. Using image enhancement techniques to improve the images visual quality help us to make better localization of pixels present in the input image which will then result in good contrast images. MRI images are low contrast images. Various methods of image enhancement help to improve the brightness and contrast of the image for practitioners to analyze and treat the infected area. After enhancement of images and identifying the tumor region, it is necessary to define the grade of the tumor. Brain tumor can be classified into two grades i.e. low grade and high grade and also four stages. In this paper we will first segment the tumor using K-means algorithm, this method will help us to extract the infected area from the body part.

The two grades of tumor can be expressed in a well defined manner. The low grade tumor is called "Benign" and the high grade tumor as "Malignant". The benign tumor is a low grade disease which does not spread over the body. Although they can be life threatening. In this the tumor cells do not grow and remain confined to a particular area but starts destroying the normal cells and tissues of that part of the brain. Therefore this grade is also a serious issue for the patient's health. On the other hand if we talk about the high grade tumors, they spread in the body part with time. These are very dangerous for patient's health and should be treated as early as possible. They spread in the brain exponentially replacing the normal cells with the infected ones by killing the tissues and veins of the brain, which eventually result in the brain to die slowly.

The extraction method will help to define the size and the shape of the tumor. Feature extraction to classify the tumor is carried out using SVM technique that may help the physicians to carry out better diagnosis.

## II. METHODOLOGY

This section aims at the techniques used for image enhancement and classification of the tumor.

### A. Histogram Equalization Techniques [3]

This technique for image enhancement deals with the histogram of the image. Histogram of the image is the plot of number of pixels for each intensity values. These methods equalize the pixels for each gray level in the input image to produce a better quality picture.

*1. Typical Histogram Equalization (HE):*

HE can be defined as, the mapping of each pixel of the input image to the relating pixels of the output. This method equalizes the intensity values to full range of the histogram to get an enhanced output. It increases the brightness and contrast of each pixel giving rise to dynamic range expansion. But, it does not consider the mean brightness of the input image into account, will gives rise to flattening of the output image histogram, false coloring, annoying artifacts in background, un-natural enhancement, excessive change in brightness, most importantly decreasing the contrast and no brightness preservation. Fig. 1 shows the input image with histogram and the output of HE with histogram.

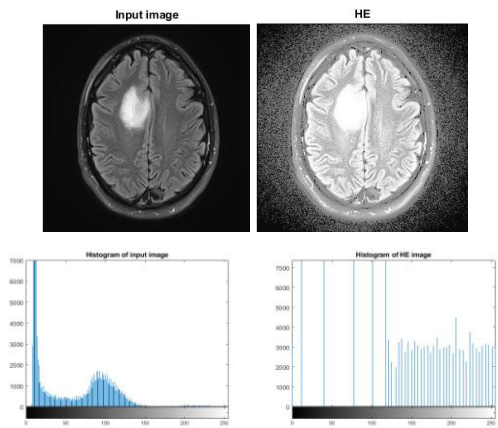

Fig. 1: Output of HE with histogram

2. *Brightness Preserving Bi-Histogram Equalization (BBHE):* [2]

BBHE algorithm is basically the upgraded version of typical HE whose main goal is to preserve brigthness and avoid false coloring. This technique partition the input picture histogram into two sub parts. The division is carried out using the average intensity of all the pixels which is said to be the input mean brightness value of all pixels that is present in the input image. After the division process using mean, these two histograms are equalized independently using the typical histogram equalization method. After performing this step, it is observed that, in the resultant image, mean brightness is present exactly between the input mean and the middle gray level. Then the two equalized images are combined together to get the resultant image. This method increases the brightness as well as the contrast of each and every pixel in a well defined manner. Using this idea it is proved that the original brightness of the input image is not lost. Fig. 2 shows the input image with histogram and the output of BBHE with histogram.

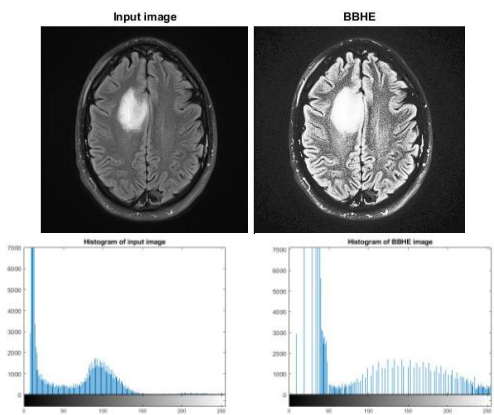

Fig. 2: Output of BBHE with histogram

3. *Recursive Mean Separated Histogram Equalization (RMSHE):*

RMSHE is an extension of the BBHE algorithm. BBHE performs mean-separation before the equalization process which helps us to preserve the images original brightness. It is already explained above, that by decomposing the input image into two sub-divisions using the mean of original picture. This result in dividing the histogram of original image, based on the mean of the histogram of original picture. But here, instead of decomposing the image once, the RMSHE algorithm decomposes the input image recursively, i.e. dividing the histogram again and again up to a scale r, till it generates 2r sub-images. The evaluated sections are equalized separately. Then all sections are combined together to give the desired result. Fig. 3 shows the input image with histogram and the output of RMSHE with histogram.

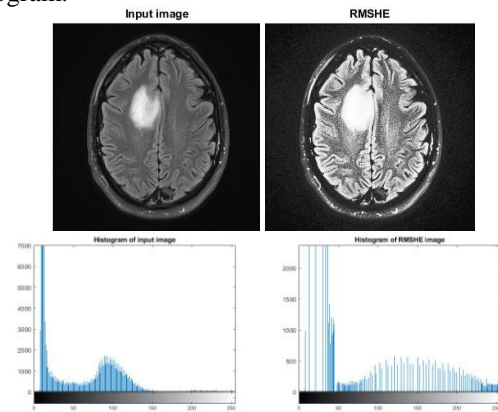

Fig. 3: output of RMSHE with histogram.

4. *Dynamic Histogram Equalization (DHE):* [4]

DHE algorithm mainly focuses on performing enhancement of the image without losing important details contained it. This method decomposes the input image histogram into many sub- sections until a new histogram is constructed that does not contain any dominating part in it. All the dynamic grey level in each of the sub-histogram is then mapped by typical HE method. All the available dynamic grey levels are divided between sub- sections on the basis of their dynamic range present in the original image and also with respect to the cumulative distributive frequency values of histograms. Separate transformation function is calculated for each sub divisions on the basis of traditional HE method having the grey levels of both original and resultant picture that are mapped. Fig. 4 shows the input image with histogram and the output of DHE with histogram.

*B. K-means Segmentation:* [7]

K-means is a method to cluster the image into K segments such that points in each cluster tend to be near to each other. It is an unsupervised method because it does not uses external classification to segment the image. Here, as we are dealing with medical images like, brain tumor image, so the picture will be segmented according to the gray level values. Area is segmented according to the center of each cluster which is the mean of data points belonging to each cluster. It is basically a method to classify or group the pixels according to the features into K groups.

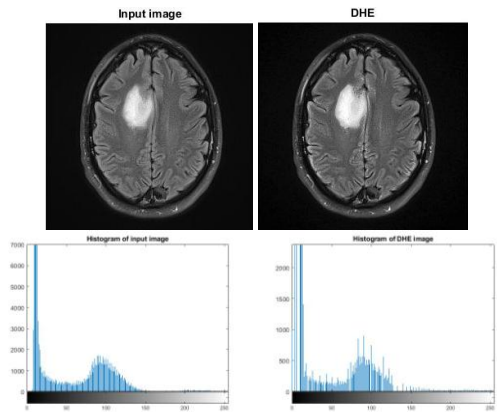

Fig. 4: Output of DHE with histogram.

This process is done to locate the Region of Interest (ROI) and its boundaries in the image. If the value of K=2 then the image will be divided into K gray levels, if K=3 then into K clusters, and so on. [5] Fig. 5 shows the input image and the K-means clustered image into K=3 and K=4.

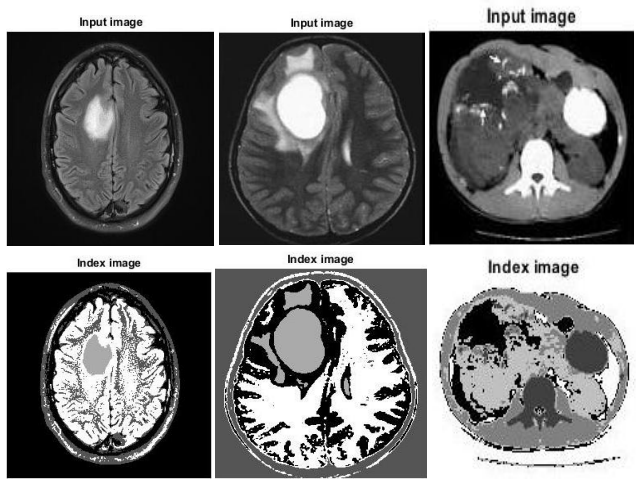

Fig. 5: Input image and the K-means clustered image into K=3 and K=4.

### C. Extration of ROI:

From the index image obtained from K-means algorithm, we can extract the ROI by identifying the K value at which the tumor is present. Selecting the correct K value, we will be able to extract only the tumor part from the brain image. We can also outline the infected part in the input image. Fig. 6 shows the extracted part of the tumor and the outlined image.

### D. Feature Extraction using SVM feature extraction: [6]

Feature extraction is done to classify the tumor into grades. Various parameters can be calculated:
1. *Shape Parameters:*
   Number of white pixels, area and perimeter.
2. *Intensity Parameters:*
   Variance, root mean square (RMS), mean, standard deviation, skewness, kurtosis and smoothness.
3. *Texture Parameters:*

Contrast, correlation, energy, homogeneity, entropy and inverse difference moment (IDM).

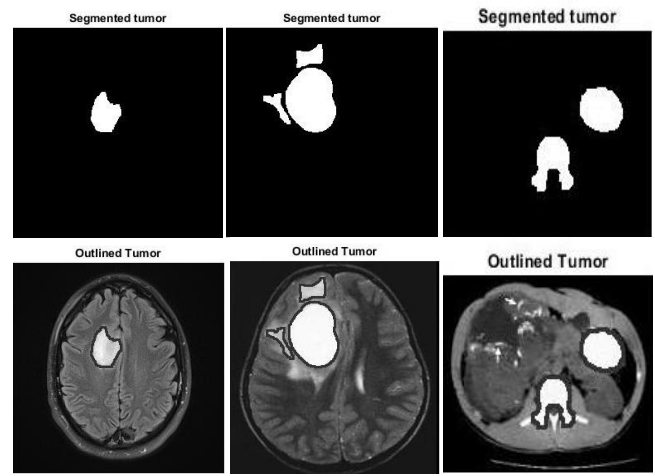

Fig. 6: Extracted part of the tumor and the outlined image.

### E. SVM Classifier trainning:

After feature extraction from the image using SVM, we can now train the classifier to classify MRI brain tumor images into low grade or high grade. Number of training datasets is used to train the classifier and make it strong. The training data should always be greater than the testing data.

### F. SVM Classification: [8]

Now as the classifier is trained to identify the grade of the tumor, we can now enter the testing images in the algorithm. Automatically the system will calculate all the different features of the input image and compare them with the training dataset. If the tumor belongs to the low grade dataset then the algorithm will define it to be benign, otherwise defining it to be malignant.

### III. FLOW CHART

Fig. 7 shows the step by step implementation of the algorithm.

### IV. RESULT

#### A. Histogram Equalization:

Following features has been extracted from different set of algorithms using test image.

1. *Peak to Signal to Noise Ratio (PSNR):*

   PSNR represents the peak error in the image. This parameter should be large, as it represents the ratio of signal power to noise power, noise power should be minimum.

2. *Mean Square Error (MSE):*

   This parameter is used to measure the average of square of error between the original image and the resultant output. MSE should minimum to get image with good quality.

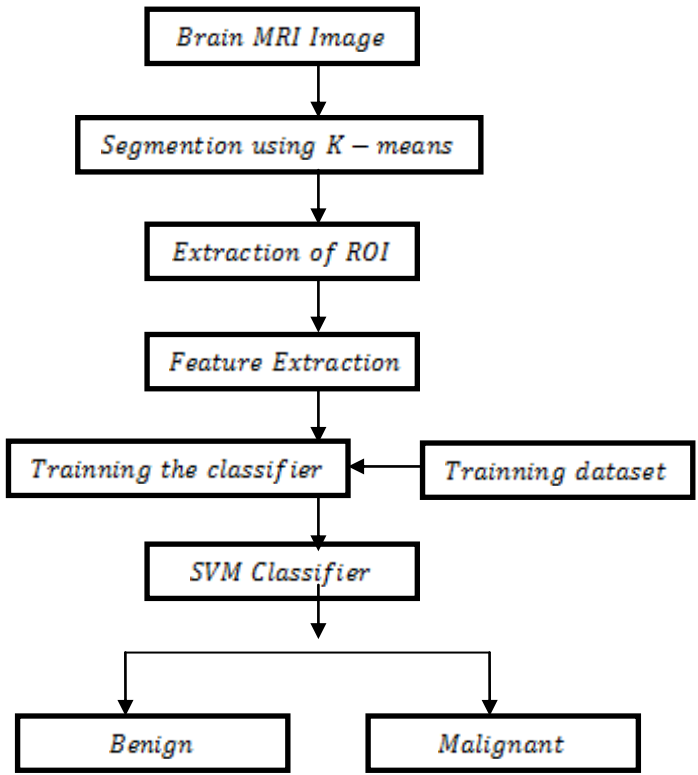

Fig.7: Flow chart

3. *Visual Quality:*

This parameter is measured by the vision of human eye, which can differentiate between best image and a moderate image. Physicians' needs images with good visual quality to identify the exact texture of the infection, regarding the same quality of the image should be good.

4. *Complexity:*

Complexity of an algorithm is calculated using the number of additions and multiplications that a particular algorithm is performing. More the complexity more time a method will take to give the output. But, if time is not the matter of constrain then the best algorithm should be selected regardless of the time taken. Table 1 shows the quality measures of multiple enhancement techniques.

TABLE 1: quality measures of multiple enhancement techniques

| Matrices | PSNR | MSE | Quality | Complexity |
|---|---|---|---|---|
| HE | 9.178198 | -2.38769 | Moderate | Very low |
| BBHE | 15.34653 | 3.780637 | Better | Low |
| RMSHE | 15.83597 | 4.270072 | Good | Moderate |
| DHE | 14.91657 | 1.68307 | Moderate | High |

B. *Segmentation and Classification:*

Following features has been extracted from the extracted tumor using test images.

1. *Shape Parameters:*

The shape parameters provide us the basic geometry of the infected part.

2. *Intensity Parameters:*

These statistical parameters tell us the hidden information about each and every pixel in an image.

3. *Texture Parameters:*

These parameters define the gray scale information of the resultant image for better analysis.

Table 2 shows the features extracted from multiple test images.

TABLE 2: features extracted from multiple test images.

| Parameters | Test Image1 | Test Image2 | Test Image3 |
|---|---|---|---|
| No. of white pixels | 1224 | 4747 | 1465 |
| Area | 426 | 1224 | 717 |
| Perimeter | 117.453 | 138.252 | 135.63 |
| Contrast | 0.0751 | 0.2237 | 0.2989 |
| Correlation | 0.9662 | 0.9584 | 0.9527 |
| Energy | 0.9617 | 0.8606 | 0.8591 |
| Homogeneity | 0.996 | 0.9987 | 0.9906 |
| Mean | 3.8808 | 16.7536 | 17.3523 |
| Standard Deviation | 28.2866 | 60.3934 | 62.907 |
| Entropy | 0.1339 | 0.3749 | 0.6839 |
| RMS | 10.465 | 33.5422 | 2.6602 |
| Variance | 0.016 | 0.0512 | 0.0576 |
| Smoothness | 1 | 1 | 1 |
| Kurtosis | 12.5456 | 53.7946 | 12.3476 |
| Skewness | 3.375 | 7.2286 | 3.3624 |
| IDM | 0.1172 | 1.7462 | 1.2871 |
| Grade | Benign | Malignant | Malignant |

V. CONCLUSION

Comparative study of various techniques is successfully performed using histogram processing. The main goal of all these equalization algorithms was to preserve the original brightness of the input image. The image should not lose its important information while performing image enhancement techniques. We observe that BBHE algorithm preserves brightness of the image to a certain extent but the improved version of BBHE is RMSHE that gives better results as observed on the basis of various parameters. Futher segmentation is carried out to extract the infected part from the MRI image. We also conclude by classifying the grade of the tumor using various features of images like shape, intensity and texture.These methods can be used by medical practitioners to perform diagnosis in a better way. Many researches have been done in the field of image processing but there is always room for improvement and future work.